\documentclass{article}
\usepackage{spconf,amsmath,graphicx}
\usepackage{cite}
\usepackage{amsmath,amssymb,amsfonts}
\usepackage{algorithmic}
\usepackage{graphicx}
\usepackage{textcomp}
\usepackage{xcolor}
\usepackage{threeparttable}
\usepackage{multirow}
\usepackage{subfigure}
\usepackage{algorithm}
\usepackage{mathrsfs}
\usepackage{colortbl}
\usepackage{multicol}
\usepackage{makecell,multirow,diagbox}
\newcommand{\tabincell}[2]{
	\begin{tabular}{@{}#1@{}}#2\end{tabular}
}

\title{Discriminatively Constrained Semi-supervised Multi-view Nonnegative Matrix Factorization with Graph Regularization}
\name{Guosheng Cui, Ruxin Wang, Dan Wu, and Ye Li$^\dagger$\thanks{$^\dagger$Corresponding author: ye.li@siat.ac.cn}}
\address{Joint Engineering Research Center for Health Big Data Intelligent Analysis Technology, \\Shenzhen Institutes of Advanced Technology, Chinese Academy of Sciences, Shenzhen, China.}
\begin{document}
%
\maketitle
\begin{abstract}
	
In recent years, semi-supervised multi-view nonnegative matrix factorization (MVNMF) algorithms have achieved promising performances for multi-view clustering. While most of semi-supervised MVNMFs have failed to effectively consider discriminative information among clusters and feature alignment from multiple views simultaneously. In this paper, a novel Discriminatively Constrained Semi-Supervised Multi-View Nonnegative Matrix Factorization (DCS$^2$MVNMF) is proposed. Specifically, a discriminative weighting matrix is introduced for the auxiliary matrix of each view, which enhances the inter-class distinction. Meanwhile, a new graph regularization is constructed with the label and geometrical information. In addition, we design a new feature scale normalization strategy to align the multiple views and complete the corresponding iterative optimization schemes. Extensive experiments conducted on several real world multi-view datasets have demonstrated the effectiveness of the proposed method.
\end{abstract}
\section{Introduction}
\label{sec:intro}
With the rapid development of the information technology, the data is not only with high dimension but also with multiple modalities or views. These different views generally contain complementary and interaction information, which can be aggregated and boost the performance for specific task in real-world applications. Consequently, the key issue is how to integrate the information came from multiple views to generate a powerful representation effectively in the learning procedure\cite{18drmvnmf,18bmvc}.

In recent years, the multi-view clustering method has obtained significant attention and application. Many nonnegative matrix factorization (NMF) based approaches for multi-view clustering have been proposed and achieved some breakthroughs\cite{17divnmf,17swdeepgmvnmf,20deepgmvnmf}. In multi-view nonnegative matrix factorization (MVNMF) framework, to integrate multiple view information into one compact representation, the key problem is to design an efficient fusion strategy. MultiNMF \cite{13multinmf} trys to learn a consensus representation with a centroid co-regularization term. In \cite{18mpmnmf}, a MVNMF approach based on pair-wise co-regularization is proposed. With pair-wise co-regularization, the features from different views are pushed together and alignment is acquired. In \cite{20udnmf}, a MVNMF method based on nonnegative matrix tri-factorization \cite{16nsnmf} is designed. This method factorizes each view into three matrices and constrain the columns of product matrix of basis matrix and shared embedding matrix to be unit vector. A centroid co-regularization is used to align the multiple views. Above mentioned methods all try to fuse multiple views in unsupervised learning manner.

Beyond that, some semi-supervised MVNMF are proposed and promising results are obtained with label information \cite{14parmvnmf,20mvocnmf,20parmvgnmf}. Wang \emph{et al.} presented the AMVNMF method, which has extended constrained nonnegative matrix factorization (CNMF) \cite{12cnmf} to handle multi-view clustering task for the first time \cite{16amvnmf}. Specifically, an auto-weighting strategy is adopted to balance different views, and a centroid co-regularization is used to align the multiple views. In \cite{19mvcnmf}, the proposed MVCNMF approach integrated the sparseness constraint and aligned multiple views with Euclidean distance based pair-wise co-regularization like \cite{18mpmnmf}. Overall, how to align multiple views and enhance the discrimination of inter-class features play an important role in feature learning.

Addressing these issues, in this paper, we present a novel discriminatively constrained semi-supervised MVNMF (DCS$^2$MVNMF), which enlarges the inter-class diversities and shrinks the intra-class variation simultaneously. In this work, a discriminative weighting matrix is designed and imposed on the auxiliary matrix to enforce the inter-class distinctness. A new graph regularization is incorporated into the objective function which has access to the geometrical information of the multi-view data in each view for decreasing the intra-class diversities. To align multiple views effectively, we restrict the column vectors of basis matrix in each view as an unit vector, and a centroid co-regularization is adopted. An iterative optimizing scheme is also designed for the proposed method.

\section{Methodology}
\label{sec:method}

\subsection{Reconstruction Loss}
In each view, we consider applying the CNMF-like reconstruction method following as \cite{16amvnmf,19mvcnmf,20mvocnmf}. Additionally, to align the multiple views, a centroid co-regularization is imposed on $\textbf{Z}^v$. Therefore, the reconstruction term of our DCS$^2$MVNMF can be written as follows:
\begin{equation}\label{reconstructTerm}
\sum\limits_{v=1}^{n_v}(||\textbf{X}^v - \textbf{W}^v ({\textbf{A}_{lc}\textbf{Z}^v} )^T||_F^2 + \gamma||{\textbf{Z}^v} - {\textbf{Z}_c}||_F^2),
\end{equation}
where $\textbf{Z}^v\in \mathbb{R}^{(n-l+c)\times d^v}$ is the auxiliary matrix for $v$th view. $\textbf{Z}_c$ indicates the consensus auxiliary matrix. $n$, $l$ and $c$ are the number of data points, the number of labeled data points and the number of classes of the datasets respectively. $d^v$ is the dimension of $v$th view. $\textbf{A}_{lc}$ is a label constraint matrix, which is as follows:

\begin{equation}\label{lcmat1}
{\textbf{A}_{lc}} = \left( {\begin{array}{*{20}{c}}
	{{{\textbf{C}}_{l \times c}}}\\
	0
	\end{array}} \right.\left. {\begin{array}{*{20}{c}}
	0\\
	{{{\textbf{I}}_{n - l}}}
	\end{array}} \right),
\end{equation}
where ${\rm{\textbf{I}}}_{n - l}$ is an identity matrix whose dimension is $n-l$. $\textbf{C}$ is a label index matrix in which each row is a one-hot vector and the $1$ corresponds to the real category.

\subsection{Discriminative Prior}
\begin{figure}[!htb]
	\begin{center}
		\includegraphics[width=0.47\textwidth]{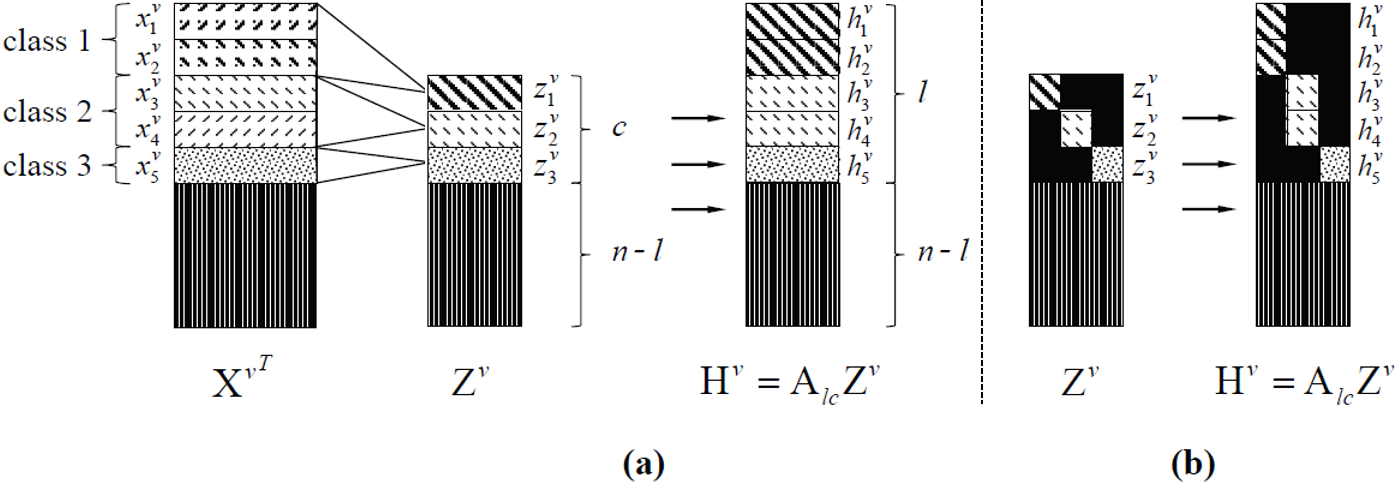}
		\setlength{\belowcaptionskip}{-0.4cm}\vspace{-0.8em}\caption{$\textbf{X}^v$, $\textbf{Z}^v$ and $\textbf{A}_{lc}\textbf{Z}^v$ of \textbf{(a)}. AMVNMF\cite{16amvnmf}, MVCNMF\cite{19mvcnmf} and MVOCNMF\cite{20mvocnmf}; \textbf{(b)}. DCS$^2$MVNMF. The samples with same labels, such as $x^v_1$ and $x^v_2$, are mapped into the same vectors, i.e., $h^v_1=h^v_2=z^v_1$.}\label{AZvAZv}
	\end{center}
\end{figure}

Only factorizing each view in CNMF framework, samples with same label are mapped to same feature vectors, which fails to consider discriminative information provided by the labeled samples and the inter-class distinctness is not guaranteed for the learned representations. (see Figure \ref{AZvAZv}). To enable the discriminative ability of DCS$^2$MVNMF, we impose following constraint on the auxiliary matrix $Z^v$:
\begin{equation}\label{ConstraintOnZv}
\sum\limits_{v=1}^{n_v}{||{\textbf{I}_{disc}} \odot {\textbf{Z}^v}||_F^2},
\end{equation}
where $\odot$ is the element-wise product operator, $\textbf{I}_{disc}=[\hat{\textbf{I}};\hat{\textbf{0}}]\in \mathbb{R}^{(n-l+c)\times d^v}$ is a discriminative matrix to ensure the distinction of the features from different classes. $\hat{\textbf{0}}\in \mathbb{R}^{(n-l)\times d^v}$ is an all zero matrix corresponding to the unlabeled samples. Specifically, $\hat{\textbf{I}}\in \mathbb{R}^{c\times d^v}$ is defined as follows:
\begin{equation}\label{hatI}
\hat{\textbf{I}} = \left( {\begin{array}{*{20}{c}}
	\bar{0}&\bar{1}& \cdots &\bar{1}\\
	\bar{1}&\bar{0}& \cdots &\bar{1}\\
	\vdots & \vdots & \ddots & \vdots \\
	\bar{1}&\bar{1}& \cdots &\bar{0}
	\end{array}} \right),
\end{equation}
where $\bar 0 = [\overbrace {0,...,0}^{{m_s}}]$ and $\bar 1 = [\overbrace {1,...,1}^{{m_s}}]$. $m_s$ is the dimensionality of each subspace. In this paper, we set $m_s=1$ like most literature. By minimizing Eq. (\ref{ConstraintOnZv}), the entries off the diagonal is suppressed, so the inter-class distinction is ensured (see Figure \ref{AZvAZv} (\textbf{b})).

\subsection{Geometrical Prior}
To further utilize the geometrical information of the multi-view, a new graph regularization  using the labeled samples and unlabeled samples is defined as follows:
\begin{equation}\label{thirdTerm}
\begin{array}{*{20}{l}}
\sum\limits_{v = 1}^{n_v} {\sum\limits_{i = 1}^n {\sum\limits_{j = 1}^n {||h_i^v - h_j^v||_2^2\textbf{S}_{ij}^v} }}  = \sum\limits_{v = 1}^{n_v} {{\rm{tr}}({\textbf{H}^v}^T{\textbf{L}^v}{\textbf{H}^v})} \\
= \sum\limits_{v = 1}^{n_v} {{\rm{tr}}({{({\textbf{A}_{lc}}{{\textbf{Z}}^v})}^T}{{\textbf{L}}^v}{\textbf{A}_{lc}}{{\textbf{Z}}^v})},
\end{array}
\end{equation}
where $\textbf{H}^v = \textbf{A}_{lc}\textbf{Z}^v$, $\textbf{L}^v=\textbf{D}^v-\textbf{S}^v$, $\textbf{D}_{ii}^v=\sum_{j}\textbf{S}_{ij}^v$ (or $\textbf{D}_{ii}^v=\sum_{j}\textbf{S}_{ji}^v$). $\textbf{S}_{ij}^v={{e^{ - ||h_i^v - h_j^v||_2^2/2{\delta ^2}}}}$, if $h^v_i$ is one of $k$-nearest-neighbors of $h^v_j$ or $h^v_j$ is one of $k$-nearest-neighbors of $h^v_i$. $\delta$ is a predefined parameter.

%
\subsection{Feature Normalization}
To align multiple features effectively, the scales of $\textbf{Z}^v$s must be restricted to be comparable. For this target, we restrict the column vectors of $\textbf{W}^v$ as $||\textbf{W}_{\cdot j}^v||_2=1$ \cite{08nge}. However, directly optimizing above objective function makes the optimization problem very complex. To tackle this problem, an alternative scheme is to compensate the norms of the basis matrix into the coefficient matrix. Then, above equation can be further rewritten as:
\begin{equation}\label{DCS2MVNMF}
\begin{array}{*{20}{l}}
{{\mathop {\min }\limits_{{\textbf{W}^v, \textbf{Z}^v,\textbf{Z}_c} \ge 0, ||\textbf{W}_{\cdot j}^v||_2=1}} \sum\limits_{v = 1}^{{n_v}} {(||{{\textbf{X}}^v} - {{\textbf{W}}^v}{{({\textbf{A}_{lc}}{{\textbf{Z}}^v})}^T}||_F^2} }\\
{ + \alpha ||{{\textbf{I}}_{disc}} \odot {{\textbf{Z}}^v}{{\textbf{Q}}^v}||_F^2} + \beta {\rm{tr}}({{\textbf{Q}}^v}^T{{({\textbf{A}_{lc}}{{\textbf{Z}}^v})}^T}{{\textbf{L}}^v}{\textbf{A}_{lc}}{{\textbf{Z}}^v}{{\textbf{Q}}^v})\\
{ + \gamma ||{{\textbf{Z}}^v}{{\textbf{Q}}^v} - {{\textbf{Z}}_c}||_F^2)},
\end{array}
\end{equation}
where $\textbf{Q}^v$ is defined as
\begin{equation}\label{definitionQv}
\begin{array}{*{20}{l}}
{{\textbf{Q}}^v} \!=\! Diag(\sqrt {\sum\limits_{i = 1}^{{m^v}} {{\textbf{W}}{{_{i,1}^v}^2}} } ,\sqrt {\sum\limits_{i = 1}^{{m^v}} {{\textbf{W}}{{_{i,2}^v}^2}} } ,...,\sqrt {\sum\limits_{i = 1}^{{m^v}} {{\textbf{W}}{{_{i,d^v}^v}^2}} } ).
\end{array}
\end{equation}

\section{Optimization}


For a specific view $v$, its optimization with $\textbf{W}^{v}$ and $\textbf{Z}^{v}$ does not depend on other views. The problem of minimizing Eq. (\ref{DCS2MVNMF}) can be written as follow:
\begin{equation}\label{dcs2mvnmfWvZv}
\begin{array}{*{20}{l}}
{\mathop {\min }\limits_{{{\textbf{W}}^v},{{\textbf{Z}}^v},{{\textbf{Z}}_c} \ge 0} ||{{\textbf{X}}^v} - {{\textbf{W}}^v}{{({\textbf{A}_{lc}}{{\textbf{Z}}^v})}^T}||_F^2} \!+\! \alpha ||{{\textbf{I}}_{disc}} \odot {{\textbf{Z}}^v}{{\textbf{Q}}^v}||_F^2\\
{ \!+\! \beta {\rm{tr}}({{\textbf{Q}}^v}^T{{({\textbf{A}_{lc}}{{\textbf{Z}}^v})}^T}{{\textbf{L}}^v}{\textbf{A}_{lc}}{{\textbf{Z}}^v}{{\textbf{Q}}^v})} \!+\! \gamma ||{{\textbf{Z}}^v}{{\textbf{Q}}^v} \!-\! {{\textbf{Z}}_c}||_F^2
\end{array}
\end{equation}


\textbf{Fixing ${\textbf{Z}_c}$, updating $\textbf{W}^v$ and $\textbf{Z}^v$}

1) Fixing ${\textbf{Z}_c}$ and $\textbf{Z}^v$, updating $\textbf{W}^v$

When ${\textbf{Z}_c}$ and $\textbf{Z}^v$ are fixed, above equation is equivalent to the following problem:
\begin{equation}\label{dcs2mvnmfWv1}
\begin{array}{*{20}{l}}
{\mathop {\min }\limits_{{{\textbf{W}}^v} \ge 0} ||{{\textbf{X}}^v} - {{\textbf{W}}^v}{{({\textbf{A}_{lc}}{{\textbf{Z}}^v})}^T}||_F^2}\\
{ + \alpha {\rm{tr}}({{\textbf{Q}}^v}^T{{({{\textbf{I}}_{disc}} \odot {{\textbf{Z}}^v})}^T}({{\textbf{I}}_{disc}} \odot {{\textbf{Z}}^v}){{\textbf{Q}}^v})}\\
{ + \beta {\rm{tr}}({{\textbf{Q}}^v}^T{{({\textbf{A}_{lc}}{{\textbf{Z}}^v})}^T}{{\textbf{L}}^v}{\textbf{A}_{lc}}{{\textbf{Z}}^v}{{\textbf{Q}}^v})}\\
{ + \gamma {\rm{tr}}({{\textbf{Q}}^v}^T{{\textbf{Z}}^v}^T{{\textbf{Z}}^v}{{\textbf{Q}}^v} - 2{{\bf{Z}}_c}^T{{\textbf{Z}}^v}{{\textbf{Q}}^v})}
\end{array}
\end{equation}

Let
\begin{equation}\label{ExY123}
\begin{array}{l}
{\textbf{Y}^v_1} = [{({\textbf{I}_{disc}} \odot {\textbf{Z}^v})^T}({\textbf{I}_{disc}} \odot {\textbf{Z}^v})] \odot \textbf{I}\\
{\textbf{Y}^v_2} = [{({\textbf{A}_{lc}}{{\textbf{Z}}^v})^T}{{\textbf{L}}^v}{\textbf{A}_{lc}}{{\textbf{Z}}^v}] \odot {\textbf{I}}
=\textbf{Y}^{v+}_2-\textbf{Y}^{v-}_2\\
{\textbf{Y}^v_3} = [{({{\textbf{Z}}^v})^T}{{\textbf{Z}}^v}] \odot {\textbf{I}}\\
{{\textbf{Y}}^v_4}={\textbf{Z}_c}^T{\textbf{Z}^v}\odot \textbf{I},
\end{array}
\end{equation}
where $\textbf{I}$ is the identity matrix and the matrices $\textbf{Y}^{v+}_2$ and $\textbf{Y}^{v-}_2$ are defined as
\begin{equation}\label{ExY2pn}
\begin{array}{l}
\textbf{Y}^{v+}_2 = [{({\textbf{A}_{lc}}{{\textbf{Z}}^v})^T}{{\textbf{D}}^v}{\textbf{A}_{lc}}{{\textbf{Z}}^v}] \odot {\textbf{I}}\\
\textbf{Y}^{v-}_2 = [{({\textbf{A}_{lc}}{{\textbf{Z}}^v})^T}{{\textbf{S}}^v}{\textbf{A}_{lc}}{{\textbf{Z}}^v}] \odot {\textbf{I}}.
\end{array}
\end{equation}
Using the Lagrange Multiplier method, we can get following update rule:
\begin{equation}\label{EXmulWv}
\begin{array}{l}
{\textbf{W}}_{ih}^v = {\textbf{W}}_{ih}^v\frac{{({{\textbf{X}}^v}{\textbf{A}_{lc}}{{\textbf{Z}}^v} + \beta {{\textbf{W}}^v}{\textbf{Y}}_2^{v - }}}{{({{\textbf{W}}^v}{({\textbf{A}_{lc}}{{\textbf{Z}}^v})}^T{\textbf{A}_{lc}}{{\textbf{Z}}^v} + \alpha {{\textbf{W}}^v}{\textbf{Y}}_1^v}}\\
\qquad\qquad \frac{{ + \gamma {{\textbf{W}}^v}{{({{\textbf{Q}}^v})}^{ - 1}}{\textbf{Y}}_4^v{)_{ih}}}}{{ + \beta {{\textbf{W}}^v}{\textbf{Y}}_2^{v + } + \gamma {{\textbf{W}}^v}{\textbf{Y}}_3^v{)_{ih}}}}.
\end{array}
\end{equation}


2) Fixing ${\textbf{Z}_c}$ and $\textbf{W}^v$, updating $\textbf{Z}^v$

After the updating of $\textbf{W}^v$, the column vectors of $\textbf{W}^v$ are normalized with $\textbf{Q}^v$ in Eq. (\ref{definitionQv}) and then the norm is conveyed to the coefficient matrix $\textbf{Z}^v$, that is:
\begin{equation}\label{normWvZv}
{{\textbf{W}}^v} \Leftarrow {{\textbf{W}}^v}{({\textbf{Q}^v})}^{ - 1},{{\textbf{Z}}^v} \Leftarrow {{\textbf{Z}}^v}{{\textbf{Q}}^v}.
\end{equation}

When ${\textbf{Z}_c}$ and $\textbf{W}^v$ are fixed, Eq. (\ref{dcs2mvnmfWvZv}) is equivalent to the following problem:
\begin{equation}\label{dcs2mvnmfZv}
\begin{array}{*{20}{l}}
{\mathop {\min }\limits_{{{\textbf{Z}}^v} \ge 0} ||{{\textbf{X}}^v} - {{\textbf{W}}^v}{{({\textbf{A}_{lc}}{{\textbf{Z}}^v})}^T}||_F^2}\\
{ + \alpha {\rm{tr}}({{({{\textbf{I}}_{disc}} \odot {{\textbf{Z}}^v})}^T}({{\textbf{I}}_{disc}} \odot {{\textbf{Z}}^v}))}\\
{ + \beta {\rm{tr}}({{({\textbf{A}_{lc}}{{\textbf{Z}}^v})}^T}{{\textbf{L}}^v}{\textbf{A}_{lc}}{{\textbf{Z}}^v})}\\
{ + \gamma {\rm{tr}}({{\textbf{Z}}^v}^T{{\textbf{Z}}^v} - 2{{\textbf{Z}}_c}^T{{\textbf{Z}}^v}).}
\end{array}
\end{equation}

%
Similarly, the update rules of ${\textbf{Z}}_{jh}^v$ can be written as following:
\begin{equation}\label{mulZv}
\begin{array}{*{20}{l}}
{{\textbf{Z}}_{jh}^v = {\textbf{Z}}_{jh}^v\frac{{({{\textbf{A}_{lc}}^T}{{\textbf{X}}^v}^T{{\textbf{W}}^v} + \beta {{\textbf{A}_{lc}}^T}{{\textbf{S}}^v}{\textbf{A}_{lc}}{{\textbf{Z}}^v}}}{{({{\textbf{A}_{lc}}^T}{\textbf{A}_{lc}}{{\textbf{Z}}^v}{{\textbf{W}}^v}^T{{\textbf{W}}^v} + \alpha {{\textbf{I}}_{disc}} \odot {{\textbf{Z}}^v}}}}\\
{\qquad \qquad \;\;{\mkern 1mu} \frac{{ + \gamma {{\textbf{Z}}_c}{)_{jh}}}}{{ + \beta {{\textbf{A}_{lc}}^T}{{\textbf{D}}^v}{\textbf{A}_{lc}}{{\textbf{Z}}^v} + \gamma {{\textbf{Z}}^v}{)_{jh}}}}.}
\end{array}\vspace{-0.3em}
\end{equation}

\textbf{Fixing $\textbf{W}^v$ and $\textbf{Z}^v$, updating ${\textbf{Z}_c}$}


Correspondingly, the exact solution for ${\textbf{Z}}_c$ is $ \frac{{\sum\limits_{v = 1}^{{n_v}} {{{\textbf{Z}}^v}} }}{{{n_v}}} \ge 0.$
\vspace{-1.0em}

\section{Experiments}

\subsection{Materials}

\textbf{Yale} \cite{yaleorl}: The dataset contains 165 gray scale images collected from 15 people.
\textbf{ORL} \cite{yaleorl}: This dataset has 400 gray scale face images collected from 40 people.
For these two datasets, the image intensity, Gabor and LBP are used as three views of the datasets.
\textbf{ECG} \footnote{https://www.physionet.org/}: This dataset has 162 original ECG records including 96 arrhythmia, 30 congestive heart failure, and 36 normal sinus rhythm records.
Each 20 seconds are treated as a data sample. In total, 294 data samples are used for evaluation. The frequency-domain and time-domain features are employed as two views.
\textbf{WebKB} \cite{webkb}: This dataset is a subset of web documents from four universities, which contains 1051 pages with 2 classes: 230 Course pages and 821 Non-Course pages. Each page has 2 views: Fulltext view and Inlinks view. This dataset is balanced by selecting first 241 data points from the second view.

\subsection{Comparative Algorithms}
To evaluate the effectiveness of the proposed method, several state-of-the-art NMF-based multi-view clustering methods are selected, including unsupervised methods (\textbf{VAGNMF}, \textbf{VCGNMF}, \textbf{LP-DiNMF} \cite{17divnmf}, \textbf{rNNMF} \cite{19rnnmf}, \textbf{MPMNMF} \cite{18mpmnmf} and \textbf{UDNMF} \cite{20udnmf}) and three semi-supervised approaches (\textbf{AMVNMF} \cite{16amvnmf}, \textbf{MVCNMF} \cite{19mvcnmf} and \textbf{MVOCNMF} \cite{20mvocnmf}). Specifically, the \textbf{VAGNMF} and the \textbf{VCGNMF} are conducted employing GNMF \cite{11gnmf} on each view individually, and use the average feature and concatenated feature of multiple views as the final representation, respectively.

To avoid errors from randomness, all methods are tested 10 times, and the average value is reported. For semi-supervised methods, 10$\%$, 20$\%$ and 30$\%$ of data points are randomly labeled for 5 times. And average result of 5 times are reported. The default parameter setting demonstrated in Table \ref{allparatable}. Two widely used evaluating metrics AC and NMI are adopted to measure the clustering performance.

\begin{table}[!htb]\addtolength{\tabcolsep}{-2.0pt}\setlength{\tabcolsep}{1.5pt}\footnotesize
	\setlength{\belowcaptionskip}{5pt}
	\begin{center}
		\vspace{-0.8em}
		\caption{Default parameter setting of DCS$^2$MVNMF. The values in brackets correspond to 10$\%$, 20$\%$ and 30$\%$ labeled data points.}\label{allparatable}
		{\begin{tabular}{c|c|c|c|c}
				\hline
				& Yale & ORL & ECG & WebKB \\
				\hline
				$\alpha$ & (10$^3$,10$^4$,10$^2$) & (10$^2$,10$^3$,10$^3$) & (10$^5$,10$^4$,10$^3$) & (10$^3$,10$^3$,10$^3$) \\
				\hline
				$\beta$ & (0.1,0.1,1) & (1,1,0.1) & (10,1,10) & (1,1,1) \\
				\hline
				$\gamma$ & (0.1,0.1,0.1) & (0.01,0.01,0.01) & (0.1,0.1,0.1) & (1,1,1) \\
				\hline
				$k$ & (2,2,2) & (2,2,2) & (4,4,4) & (3,3,3) \\
				\hline
		\end{tabular}}
	\end{center}\vspace{-1.5em}
\end{table}

\begin{table*}[!htb]\addtolength{\tabcolsep}{-2.0pt}\setlength{\tabcolsep}{1.5pt}\footnotesize
	\setlength{\belowcaptionskip}{5pt}
	\begin{center}
		\vspace{-1.2em}
		\caption{Clustering performance on four datasets compared with state-of-the-art unsupervised methods. (note: \textbf{MPMNMF$\_$1} and \textbf{MPMNMF$\_$2} adopt the Euclidean distance based and kernel based pair-wise co-regularization, respectively.)}\label{allunsuptable}
		{\begin{tabular}{c|c|c|c|c|c|c|c|c|c}
				\hline
				Datasets & Metrics & VAGNMF & VCGNMF & LP-DiNMF & rNNMF & MPMNMF$\_$1 & MPMNMF$\_$2 & UDNMF & \tabincell{c}{DCS$^2$MVNMF} \\ 
				\hline
				\multirow{2}{*}{Yale}  & AC & 48.55$\pm$2.95 & 53.52$\pm$3.14 & 51.82$\pm$3.03 & 51.15$\pm$5.49 & 50.00$\pm$3.05 & 49.52$\pm$3.67 & 47.76$\pm$2.83 & \textbf{61.03$\pm$3.57} \\
				& NMI & 52.38$\pm$2.70 & 56.79$\pm$2.27 & 53.73$\pm$2.67 & 53.00$\pm$4.40 & 52.89$\pm$2.78 & 54.00$\pm$2.57 & 49.88$\pm$2.73 & \textbf{63.35$\pm$2.18} \\
				\hline
				\multirow{2}{*}{ORL}  & AC & 68.78$\pm$3.17 & 70.28$\pm$2.26 & 68.48$\pm$3.33 & 64.35$\pm$2.74 & 70.25$\pm$2.87 & 69.48$\pm$2.89 & 61.73$\pm$1.56 & \textbf{76.86$\pm$1.46} \\
				& NMI & 82.88$\pm$1.68 & 83.67$\pm$1.01 & 82.78$\pm$1.80 & 79.79$\pm$1.31 & 83.65$\pm$0.99 & 83.64$\pm$1.31 & 78.54$\pm$1.03 & \textbf{87.09$\pm$0.58} \\
				\hline
				\multirow{2}{*}{ECG}  & AC & 55.41$\pm$4.58 & 56.43$\pm$2.93 & 54.90$\pm$5.69 & 57.38$\pm$0.45 & 57.86$\pm$3.60 & 58.20$\pm$3.71 & 56.84$\pm$3.11 & \textbf{70.63$\pm$6.19} \\
				& NMI & 22.82$\pm$4.34 & 26.36$\pm$4.43 & 22.52$\pm$7.80 & 23.80$\pm$1.24 & 19.72$\pm$4.60 & 20.41$\pm$4.42 & 22.28$\pm$3.25 & \textbf{40.52$\pm$6.79} \\
				\hline
				\multirow{2}{*}{WebKB}  & AC & 77.02$\pm$1.24 & 79.57$\pm$0.70 & 76.49$\pm$0.79 & 78.30$\pm$1.17 & 81.11$\pm$1.77 & 78.98$\pm$2.10 & 79.34$\pm$4.80 & \textbf{92.09$\pm$1.46} \\
				& NMI & 26.84$\pm$2.28 & 29.42$\pm$1.42 & 26.14$\pm$1.56 & 24.63$\pm$5.85 & 34.22$\pm$2.46 & 30.99$\pm$5.15 & 29.98$\pm$9.00 & \textbf{63.64$\pm$5.18} \\
				\hline
		\end{tabular}}
	\end{center}\vspace{-1.5em}

\end{table*}

\begin{table*}[!htb]\addtolength{\tabcolsep}{-2.0pt}\setlength{\tabcolsep}{1.5pt}\footnotesize
	\setlength{\belowcaptionskip}{5pt}
	\begin{center}
		\vspace{-1.2em}
		\caption{Clustering performance on four datasets compared with state-of-the-art semi-supervised methods.}\label{allsuptable}
		{\begin{tabular}{c|c|c|c|c|c|c|c|c|c}
				\hline
				&  & \multicolumn{2}{c}{AMVNMF}&\multicolumn{2}{|c}{MVCNMF}&\multicolumn{2}{|c}{MVOCNMF}&\multicolumn{2}{|c}{DCS$^2$MVNMF}\\
				\hline
				&  & AC & NMI & AC & NMI & AC & NMI & AC & NMI \\
				\hline
				\multirow{3}{*}{Yale}
				&10$\%$ & 50.40$\pm$0.89 & 53.51$\pm$0.75 & 50.99$\pm$1.53 & 54.54$\pm$1.59 & 51.33$\pm$1.62 & 54.61$\pm$1.10 & \textbf{61.03$\pm$3.57} & \textbf{63.35$\pm$2.18} \\
				&20$\%$ & 55.38$\pm$1.97 & 59.08$\pm$2.04 & 57.68$\pm$1.12 & 61.28$\pm$1.24 & 57.84$\pm$1.44 & 61.34$\pm$0.90 & \textbf{70.88$\pm$3.31} & \textbf{69.02$\pm$2.47} \\
				&30$\%$ & 61.56$\pm$1.70 & 65.45$\pm$1.68 & 62.63$\pm$1.56 & 67.00$\pm$1.02 & 63.70$\pm$2.63 & 67.57$\pm$1.40 & \textbf{77.50$\pm$2.06} & \textbf{74.92$\pm$2.43} \\
				\hline
				\multirow{3}{*}{ORL}
				&10$\%$ & 63.15$\pm$0.65 & 79.16$\pm$0.51 & 66.48$\pm$0.80 & 81.41$\pm$0.40 & 66.77$\pm$1.29 & 81.22$\pm$0.56 & \textbf{76.86$\pm$1.46} & \textbf{87.09$\pm$0.58} \\
				&20$\%$ & 68.41$\pm$0.54 & 82.37$\pm$0.29 & 70.18$\pm$0.85 & 83.76$\pm$0.28 & 70.27$\pm$0.55 & 83.81$\pm$0.40 & \textbf{85.69$\pm$3.02} & \textbf{91.58$\pm$1.38} \\
				&30$\%$ & 71.99$\pm$1.20 & 84.51$\pm$0.42 & 74.02$\pm$1.41 & 85.98$\pm$0.50 & 75.27$\pm$0.54 & 86.55$\pm$0.13 & \textbf{89.13$\pm$1.47} & \textbf{92.68$\pm$1.03} \\
				\hline
				\multirow{3}{*}{ECG}
				&10$\%$ & 50.90$\pm$0.96 & 13.90$\pm$1.19 & 53.78$\pm$0.56 & 26.23$\pm$0.87 & 58.69$\pm$1.15 & 26.31$\pm$1.39 & \textbf{70.63$\pm$6.19} & \textbf{40.52$\pm$6.79} \\
				&20$\%$ & 54.81$\pm$2.34 & 18.93$\pm$1.51 & 57.50$\pm$1.57 & 28.86$\pm$0.89 & 63.74$\pm$1.84 & 25.51$\pm$1.58 & \textbf{82.26$\pm$3.25} & \textbf{56.08$\pm$6.23} \\
				&30$\%$ & 55.67$\pm$3.63 & 19.20$\pm$3.54 & 66.22$\pm$1.49 & 31.46$\pm$1.24 & 70.29$\pm$2.65 & 36.43$\pm$2.16 & \textbf{84.97$\pm$1.24} & \textbf{58.45$\pm$2.62} \\
				\hline
				\multirow{3}{*}{WebKB}
				&10$\%$ & 82.44$\pm$1.54 & 33.26$\pm$3.55 & 82.73$\pm$2.39 & 38.54$\pm$4.39 & 83.29$\pm$1.96 & 39.89$\pm$3.84 & \textbf{92.09$\pm$1.46} & \textbf{63.64$\pm$5.18} \\
				&20$\%$ & 85.90$\pm$1.94 & 41.58$\pm$5.12 & 84.44$\pm$2.05 & 43.13$\pm$3.01 & 81.87$\pm$2.61 & 39.35$\pm$3.54 & \textbf{92.35$\pm$2.58} & \textbf{62.12$\pm$8.44} \\
				&30$\%$ & 90.17$\pm$1.00 & 53.84$\pm$3.19 & 91.06$\pm$0.73 & 59.09$\pm$1.66 & 90.45$\pm$0.68 & 57.14$\pm$1.58 & \textbf{95.03$\pm$1.68} & \textbf{71.92$\pm$7.23} \\
				\hline
		\end{tabular}}
	\end{center}\vspace{-1.5em}
\end{table*}

\begin{figure}[!t]
	\centerline{\includegraphics[width=0.45\textwidth]{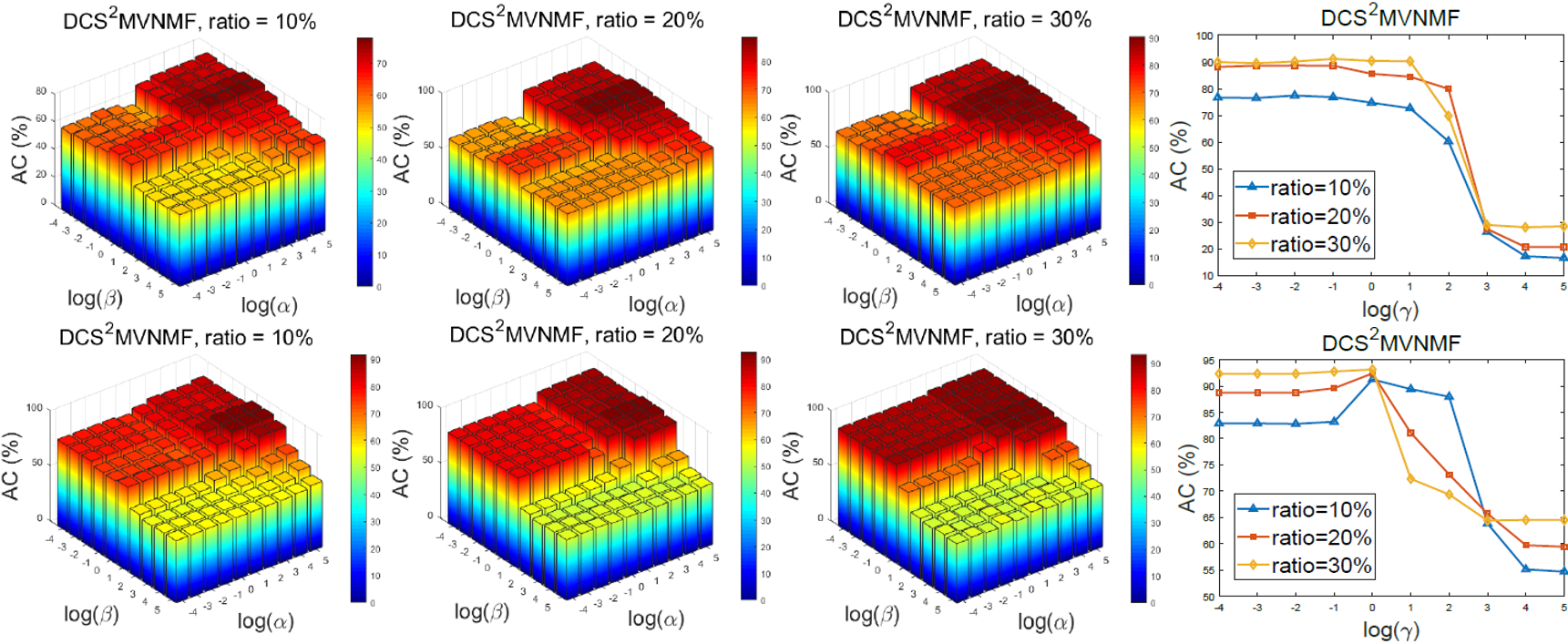}}
	\caption{DCS$^2$MVNMF with different parameters $\alpha$, $\beta$ and $\gamma$ on ORL (first row) and WebKB (second row) datasets.}
	\label{fig:2}
\end{figure}

\begin{table}[!htb]\addtolength{\tabcolsep}{-2.0pt}\setlength{\tabcolsep}{1.5pt}\footnotesize
	\setlength{\belowcaptionskip}{5pt}
	\begin{center}
		\vspace{-1.2em}
		\caption{Ablation study on Yale, ORL datasets with 10$\%$ labeled data points.}\label{t3}
		{\begin{tabular}{l|c|c|c|c}
				\hline
				&  \multicolumn{2}{c}{Yale} & \multicolumn{2}{c}{ORL}\\
				\hline
				&  AC & NMI & AC & NMI \\
				\hline
				Baseline & \ \  46.16 \ \   & \ \  50.40 \ \   & \ \  60.13 \ \   & \ \  76.70 \ \   \\
				Baseline$+\alpha$ & 56.76 & 59.02 & 71.14 & 82.61 \\
				Baseline$+\beta$ & 50.10 & 53.56 & 67.70 & 83.19 \\
				DCS$^2$MVNMF w/o $\textbf{Q}^v$ & 59.25 & 61.75 & 70.31 & 83.79 \\
				DCS$^2$MVNMF & \textbf{61.03} & \textbf{63.35} & \textbf{76.86} & \textbf{87.09} \\
				\hline
		\end{tabular}}
	\end{center}\vspace{-1.5em}
\end{table}

\subsection{Comparison Results}
Table \ref{allunsuptable} shows the comparison results of some recently proposed unsupervised MVNMFs and our DCS$^2$MVNMF with 10$\%$ labels on four datasets. From this table, with the utilization of label information the clustering performance is effectively improved. Table \ref{allsuptable} lists the similar comparison results with three CNMF based semi-supervised MVNMFs using different labeling ratios. It demonstrates that the discriminative information is effectively explored in our DCS$^2$MVNMF. And the introducing of discriminative prior and graph regularization can help to enhance the inter-class distinction and shrink the intra-class variation simultaneously.

\subsection{Parameter Analysis}


We also conduct the parameter analysis on ORL and WebKB datasets. Figure \ref{fig:2} shows the performance of semi-supervised DCS$^2$MVNMF on ORL and WebKB datasets, with different ratios of labeled data points (10$\%$, 20$\%$, 30$\%$ respectively), versus parameters $\alpha$, $\beta$ and $\gamma$. We can see that when $\alpha < 10^2$ and $\beta > 10^0$ with $\gamma \le 10^1$ or $\gamma \le 10^0$, the performance of the proposed method is better.

\subsection{Ablation Analysis}

Table \ref{t3} shows the ablation results of our method on Yale and ORL datasets. In the ablation study, the reconstruction term with consensus align term are choose as the ``Baseline''. From the table, it shows that the discriminative prior (``Baseline$+\alpha$'') and the geometrical prior (``Baseline$+\beta$'') effectively boost the performance of clustering with considerable improvement on these two datasets. In addition, it also suggests that the proposed feature normalization is beneficial to multiple feature alignment and simplification optimization.

\section{Conclusion}
In this paper, a novel semi-supervised MVNMF with discriminative and geometrical prior is proposed, which enlarges the inter-class diversities and shrinks the intra-class variation simultaneously. In order to effectively align the features of multiple views, a new feature normalizing scheme is adopted to restrict the feature scales of different views. Experiments on four datasets have verified the effectiveness of our method.

\vfill\pagebreak
%

\newpage
\bibliographystyle{IEEEbib}
\bibliography{Article-2021}

\end{document}